\documentclass[10pt, a4paper]{article}

\usepackage[final]{lrec2026} 

\usepackage{times}
\usepackage{latexsym}
\usepackage{subfigure}
\usepackage{float}
\usepackage{amsmath}
\usepackage{xcolor}

\usepackage[T1]{fontenc}
\usepackage[utf8]{inputenc}

\usepackage{microtype}

\usepackage{inconsolata}

\usepackage{graphicx}
\graphicspath{{images/}}
\usepackage{multirow}
\usepackage{booktabs}  

\title{SlovKE: A Large-Scale Dataset and LLM Evaluation for Slovak Keyphrase Extraction}

\name{David Števaňák\textsuperscript{1,4} \quad Marek Šuppa\textsuperscript{2,3,4}}

\address{
  \textsuperscript{1}University of Vienna, Austria\\
  \textsuperscript{2}Faculty of Mathematics, Physics and Informatics, Comenius University in Bratislava\\
  \textsuperscript{3}Cisco Systems, Inc.~\textsuperscript{4}NaiveNeuron\\
  \textsuperscript{\textbf{Correspondence:}~\texttt{marek@suppa.sk}}
}

\abstract{
Keyphrase extraction for morphologically rich, low-resource languages remains understudied, largely due to the scarcity of suitable evaluation datasets. We address this gap for Slovak by constructing a dataset of 227,432 scientific abstracts with author-assigned keyphrases---scraped and systematically cleaned from the Slovak Central Register of Theses---representing a 25-fold increase over the largest prior Slovak resource and approaching the scale of established English benchmarks such as KP20K. Using this dataset, we benchmark three unsupervised baselines (YAKE, TextRank, KeyBERT with SlovakBERT embeddings) and evaluate KeyLLM, an LLM-based extraction method using GPT-3.5-turbo. Unsupervised baselines achieve at most 11.6\% exact-match $F1@6$, with a large gap to partial matching (up to 51.5\%), reflecting the difficulty of matching inflected surface forms to author-assigned keyphrases. KeyLLM narrows this exact--partial gap, producing keyphrases closer to the canonical forms assigned by authors, while manual evaluation on 100 documents ($\kappa = 0.61$) confirms that KeyLLM captures relevant concepts that automated exact matching underestimates. Our analysis identifies morphological mismatch as the dominant failure mode for statistical methods---a finding relevant to other inflected languages. The dataset (\url{https://huggingface.co/datasets/NaiveNeuron/SlovKE}) and evaluation code (\url{https://github.com/NaiveNeuron/SlovKE}) are publicly available.
\\ \newline \Keywords{keyphrase extraction, Slovak, morphologically rich languages, large language models, benchmark dataset, evaluation metrics, Slavic NLP} }

\begin{document}

\maketitleabstract

\section{Introduction}

Keyphrase extraction---identifying the words or phrases that best represent a document's topics \citep{hasan-ng-2014-automatic}---is essential for scientific literature discovery and classification \citep{papagiannopoulou2020review}. While recent advances have yielded strong results for English and other high-resource languages \citep{alzaidy2019bi, basaldella2018bidirectional, nguyen2007keyphrase}, morphologically rich low-resource languages remain underexplored. In such languages, a single lemma can surface in dozens of inflected forms \citep{tsarfaty2013parsing}, creating a fundamental mismatch between the surface forms models extract and the canonical forms authors assign as keyphrases. This challenge is not unique to any single language---it affects the entire family of morphologically rich languages, including most Slavic, Finno-Ugric, and Turkic languages \citep{under-resourced-lang}. Figure~\ref{fig:orava_marketing} illustrates the problem: author-assigned keyphrases in nominative form (e.g., \textit{Rozvojový potenciál}) appear in the abstract only in inflected variants (e.g., genitive \textit{rozvojového potenciálu}).

We study this problem through the lens of Slovak, a West Slavic language with limited keyphrase extraction research due to the lack of suitable corpora. Building on the work of \citet{Zelinka2023}, who collected approximately 9,000 documents but faced limitations in data quality and scale, we construct a substantially larger and cleaner dataset and conduct a systematic evaluation of both established and LLM-based keyphrase extraction methods. We evaluate KeyLLM \citep{keyllm}, an LLM-based extraction approach, to determine whether generative models can better handle the morphological variability that hinders traditional extractive methods.

\begin{figure*}[ht]
    \centering
    \fbox{\parbox{0.95\textwidth}{ 
        \raggedright 
        Predmetom skúmania diplomovej práce je potenciál \textbf{\textcolor{green}{vidieka}}, územný rozvoj a možnosti \textbf{\textcolor{teal}{rozvoja vidieka}}. Objektom skúmania je Žilinský samosprávny kraj a jeho historický \textcolor{blue}{región} Dolná Orava. Cieľom práce je na základe teoretického rozpracovania problematiky a analýzy \textbf{\textcolor{red}{rozvojového potenciálu}} vymedziť možnosti \textbf{\textcolor{teal}{rozvoja vidieka}}. V práci je sústredená pozornosť na analýzu \textbf{\textcolor{red}{rozvojového potenciálu}} Žilinského kraja a možností využitia jeho jednotlivých zložiek na zabezpečenie ekonomického, sociálneho a trvaloudržateľného \textbf{\textcolor{teal}{rozvoja}}. Predpokladaný prínos práce vidíme vo formulovaní návrhov využitia \textbf{\textcolor{red}{rozvojového potenciálu}} kraja na zabezpečenie endogénneho \textbf{\textcolor{teal}{rozvoja vidieka}}.

        \vspace{1em}
        \textbf{\textit{Keywords}}:  
        \textbf{\textcolor{green}{Vidiek}}, \textbf{\textcolor{teal}{Rozvoj vidieka}}, \textbf{\textcolor{red}{Rozvojový potenciál}}, Vidiecky \textcolor{blue}{región}, Vidiecke sídlo, Nástroje rozvoja
    }}
    \caption{Example abstract from the Test22K dataset (Slovak). Color coding indicates keyphrase occurrences: author-assigned keyphrases appear in the text in various inflected forms (\textcolor{red}{red}: exact match, \textcolor{teal}{teal}: partial overlap, \textcolor{green}{green}: single-word match, \textcolor{blue}{blue}: shared fragment). Note that surface forms in the abstract (e.g., \textit{rozvojového potenciálu}, genitive) differ from the canonical keyphrase form (\textit{Rozvojový potenciál}, nominative), illustrating the morphological mismatch challenge.}
    \label{fig:orava_marketing}
\end{figure*}

Our main contributions are:

\begin{itemize}
    \item A rigorously cleaned dataset of 227,432 Slovak scientific abstracts with author-assigned keyphrases---a 25-fold increase over prior work---whose size and statistics are comparable to established English benchmarks such as KP20K \citep{meng2017deep}.
    \item A controlled evaluation of statistical (YAKE), graph-based (TextRank), and embedding-based (KeyBERT) models on Slovak, revealing that exact-match $F1@6$ remains below 12\% while partial-match $F1@6$ reaches 51.5\%, exposing a ${\sim}$40-point gap attributable to morphological inflection.
    \item Evidence that KeyLLM substantially narrows this exact--partial gap by generating keyphrases in canonical form rather than extracting surface tokens, suggesting that generative models are better suited to morphologically rich languages.
    \item A manual evaluation and error analysis ($\kappa{=}0.61$) identifying morphological mismatch as the dominant failure mode for extractive models and unmotivated adjective extraction as the primary weakness of KeyLLM.
\end{itemize}

\section{Related Work}

Keyphrase extraction has evolved from early statistical and graph-based methods \citep{hulth2003improved, mihalcea2004textrank} to modern transformer-based approaches. Recent advances leveraging pre-trained language models (PLMs) have achieved state-of-the-art performance \citep{song2023survey}, with BERT-based methods like KeyBERT \citep{grootendorst2020keybert} and approaches utilizing domain-specific models like SciBERT \citep{park2020scientific} showing significant improvements. Supervised methods formulating keyphrase extraction as sequence labeling with BiLSTM-CRF architectures over contextualized embeddings \citep{sahrawat2020keyphrase} have demonstrated strong performance, though they require substantial training data. More recently, large language models have been explored for keyphrase extraction \citep{keyllm}, though primarily for high-resource languages. \citet{wu2022pretrained} demonstrated that PLMs exhibit significant performance degradation on low-resource languages, highlighting the challenges for languages like Slovak.

This performance gap has motivated a growing body of work on low-resource and multilingual keyphrase extraction. \citet{gao2022retrieval} proposed retrieval-augmented multilingual keyphrase generation, while multilingual datasets like MAKED \citep{kulkarni2022maked}, TermEval \citep{terryn2019termeval}, and the recent EUROPA dataset \citep{salaun2024europa} covering 24 European languages support cross-lingual research. However, most multilingual work focuses on Romance and Germanic languages, with limited exploration of morphologically rich Slavic languages. Morphological complexity presents unique challenges \citep{tsarfaty2013parsing}, as languages like Slovak exhibit extensive inflection, with a single lemma having dozens of surface forms, complicating keyphrase matching and evaluation. Critically, no prior work has systematically quantified the impact of this inflectional variation on standard evaluation metrics---a gap our study addresses by reporting both exact and partial matching and analyzing the divergence between them.

Among Slavic languages, Slovak, Czech, and Polish are considered \textit{under-resourced}, with only partial NLP infrastructure available (e.g., corpora, taggers, morphological analyzers) \citep{under-resourced-lang}. Within the Slavic family, Czech has seen more development, with \citet{docekal2022query} testing a query-based method on 48,879 documents, though performance worsened on abstracts. Polish keyphrase extraction has also been studied \citep{giarelis2021comparative, pkezik2022keyword}, with mixed results for deep learning approaches.

Slovak keyphrase extraction is still in its early stages. \citet{varga2022keyphrase} applied unsupervised methods to five court decisions, with TF-IDF achieving the best results. \citet{Zelinka2023} made an important first step by collecting approximately 9,000 documents from the Slovak thesis registry; however, the dataset inherits noise typical of large-scale web harvesting---mixed-language content, inconsistent metadata, and variable keyphrase formatting---challenges that motivate the more extensive cleaning pipeline we describe in Section~\ref{sec:dataset_cleaning}.

The main datasets for benchmarking keyphrase extraction in English include KPTimes \citep{gallina2019kptimes} (news articles) and, for scientific articles, KP20K \citep{meng2017deep}, Inspec \citep{hulth2003improved}, and SemEval \citep{kim-etal-2010-semeval}. The KP20K dataset, with 530,000 training and 20,000 test documents, is most relevant to our work. While English benefits from hundreds of thousands of annotated documents, Slavic languages rely on datasets orders of magnitude smaller, making it difficult to determine whether low extraction performance reflects genuine model limitations or merely insufficient data.

\section{Methodology}

We select baselines that span three dominant paradigms---statistical (YAKE), graph-based (TextRank), and embedding-based (KeyBERT)---following the configuration of \citet{Zelinka2023} to enable direct comparison with prior Slovak work. Crucially, all three are \emph{extractive}: they return surface tokens from the input text, making them structurally vulnerable to the morphological mismatch between in-text inflected forms and author-assigned canonical keyphrases. This property is not specific to Slovak; it affects every morphologically rich language and makes these baselines a meaningful diagnostic tool for quantifying the impact of inflection on evaluation metrics. We contrast them with KeyLLM, a large-language-model-based approach that \emph{generates} keyphrases and is therefore free to produce canonical forms regardless of the surface forms present in the text.

\subsection{YAKE}

YAKE \citep{campos2020yake} is an unsupervised keyword extraction algorithm that uses statistical methods to analyze document words. It scores keyphrases based on factors such as casing, term position, frequency, and relatedness to context. We use the \texttt{yake} library with Slovak stopwords and a uni- to bi-gram n-gram range, consistent with prior work.

\subsection{TextRank}

TextRank \citep{mihalcea2004textrank} is another unsupervised approach that employs a graph-based structure, where keyphrases form nodes in a graph, and their relevance is computed iteratively using a variant of the PageRank algorithm \citep{rogers2002google}. The \texttt{pke} library is used with custom adjustments to favor uni- and bi-grams.

\subsection{KeyBERT}

KeyBERT \citep{grootendorst2020keybert} leverages BERT embeddings to extract keyphrases by measuring their cosine similarity to the entire document. The approach assumes that a word effectively describing the document will have a vector representation close to that of the document itself. In our approach, we utilize the \href{https://huggingface.co/kinit/slovakbert-sts-stsb}{\texttt{kinit/slovakbert-sts-stsb}} \citep{pikuliak-etal-2022-slovakbert}, which has been fine-tuned for Slovak, and apply it to generate uni- and bi-gram keyphrases. 

\subsection{KeyLLM}

KeyLLM builds upon KeyBERT and leverages Large Language Models (LLMs) such as GPT \citep{radford2018improving} for keyphrase extraction. Unlike KeyBERT, which relies on cosine similarity between BERT embeddings, KeyLLM uses prompted LLMs (GPT-3.5 \citep{ouyang2022training} or GPT-4 \citep{achiam2023gpt}) to directly generate keyphrases from the document text.

To reduce computational costs, KeyLLM offers an optional embedding-based clustering approach where documents are encoded using a sentence transformer (\texttt{all-MiniLM-L6-v2} for English or \texttt{kinit/slovakbert-sts-stsb} for Slovak). A similarity threshold parameter determines clustering: higher thresholds (0.90) create smaller clusters requiring more LLM calls, while lower thresholds (0.75) reduce costs by grouping diverse documents together. We analyze both versions---with clustering at various thresholds (0.75, 0.85, 0.90) and without embeddings---representing the first application to Slovak.

\section{Dataset}

\subsection{Data Collection}

Our data source is the Central Register of Theses and Dissertations in Slovakia (\href{https://opac.crzp.sk/}{https://opac.crzp.sk/}). It is a publicly available database of bachelor, diploma, rigorous, and similar types of theses from Slovak universities.

\begin{figure}[htbp]
    \centering
    \includegraphics[width=.45\textwidth]{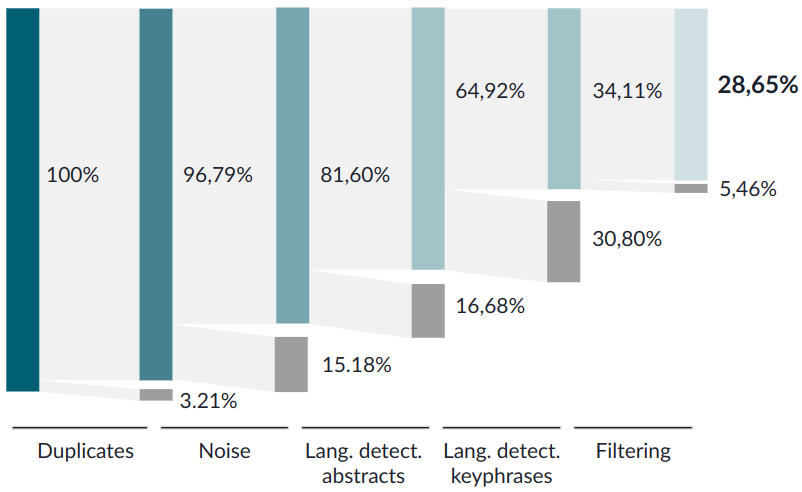}
    \caption{Filtering process and percentage of rows in the original dataset}
    \label{fig:filt_proces}
\end{figure}

The theses have been stored since 2010, which provided us with a substantial number of available documents. At the time of our data collection (2024), there were a total of 794,527 available documents. Each thesis record in the register includes what the system labels a \textit{primary abstract} (written in the thesis's main language, typically Slovak) and a \textit{secondary abstract} (a translation, typically into English), along with author-assigned keyphrases and metadata such as university, year of completion, and supervisor.

The website's session limit of 20,000 documents required strategic filtering by year and alphabetical prefixes to access nearly all records through parallel processing. We successfully retrieved 793,722 documents, with the remaining 805 excluded due to non-alphabetical titles or unreadable text.

\subsection{Data Cleaning}
\label{sec:dataset_cleaning}

Starting from 793,722 scraped records, we applied a multi-stage cleaning pipeline (detailed in Appendix~\ref{sec:cleaning_details}) consisting of: (1)~duplicate removal, prioritizing records with complete abstracts and keyphrases; (2)~recovery of keyphrases appended to abstract text by universities lacking a dedicated keyphrase field; (3)~removal of noisy metadata prepended to abstracts (e.g., author names, thesis type, page counts); (4)~language verification using the \texttt{lingua} library \citep{lingua2021}, which revealed that 20\% of Slovak-labeled abstracts were actually in English; (5)~keyphrase normalization using \texttt{Stanza} POS tagging \citep{qi2020stanza} to split concatenated keyphrase lists and enforce a maximum length of four words; and (6)~length-based filtering, retaining abstracts of 500--2000 characters with 4--15 keyphrases. The overall filtering process is summarized in Figure~\ref{fig:filt_proces}.

\subsection{Dataset Statistics}

\begin{table*}
  \centering
  \begin{tabular}{llllll}
          \hline
        \textbf{Dataset (Test)} & \textbf{Inspec} & \textbf{KP20K} & \textbf{SemEval} & \textbf{Zelinka} & \textbf{SlovKE} \\
        \hline
        \textbf{Annotator} & Expert & Author  & Author+Reader & Author & Author \\
        \textbf{Document Type} & Abstracts & Abstract &  Papers (6-8p)  & Abstracts & Abstracts  \\
        \textbf{\# Documents (Train)} & 1000 & 530K & 144 & - & 182K \\
        \textbf{\# Documents (Validation)} & 500 & 20K & - & - & 22K \\
        \textbf{\# Documents (Test)} & 500 & 20K & 100 & 9K & 22K \\
        \hline
        \textbf{Avg. no. Words per Document} & 134.6 & 176 & 7,961 &  125.63 & 134.02 \\
        \textbf{Avg. no. of Keyphrases} & 9.8 & 5.3 & 14.7 & 6.08 & 5 \\
        \textbf{Avg. len of Keyphrase} & 2.3 & 2.6 & 2.2 & 1.68 & 1.68  \\
        \textbf{\% of absent Keyphrases} & 44.31 &  42.6 & 19.7 & 49.1 & 37 \\
        \hline
    \end{tabular}
    \caption{Dataset comparison for Test sets}
    \label{tab:datasets_comparison}
\end{table*}

After cleaning, we obtained 227,432 records, which we release
as the \textbf{SlovKE} (Slovak Keyphrase Extraction) dataset.
We split them randomly (with a fixed seed and without explicit
stratification) into training (80\%), validation (10\%), and
test (10\%) sets. We additionally created a larger 20\% test
split (Test45K) for preliminary analysis; since no significant
differences were found between the two test splits, we report
results only for the 10\% split (Test22K). The Test45K
documents were not added back to the training set. As all
evaluated models are unsupervised, the training split was
not used for model training; it is included to support
future supervised approaches.

The dataset statistics, compared to other scholarly datasets, are presented in Table \ref{tab:datasets_comparison}. As shown, our dataset aligns well in size with larger keyphrase extraction datasets in English, even in terms of statistical properties. Notably, our absent-keyphrase rate (37\%) is comparable to that of KP20K (42.6\%) and Inspec (44.3\%). This indicates that the challenge of absent keyphrases in Slovak is not disproportionately harder than in English---contrary to what might be expected for a morphologically rich language---and that models and evaluation protocols developed for English absent-keyphrase generation can be meaningfully applied cross-lingually. As pointed out by \citet{meng2017deep}, a substantial absent-keyphrase rate is not undesirable; it can enhance model training and evaluation by testing the model's ability to generate keyphrases that are not explicitly present in the text. The keyphrases predominantly consist of unigrams and bigrams.

It is worth noting that author-assigned keyphrases are inherently subjective: different authors may select different terms for similar content, and individual variation in keyphrase granularity, abstraction level, and terminology is well documented even in English datasets \citep{kim-etal-2010-semeval}. In our dataset, this subjectivity is compounded by the fact that keyphrases are assigned by students with varying levels of academic experience, without standardized guidelines across universities. Rather than treating this as a limitation, we view it as representative of the realistic annotation conditions under which keyphrase extraction systems must operate, particularly in low-resource settings where expert re-annotation is impractical.

Additional statistics on abstract length distribution and token counts are provided in Appendix~\ref{sec:additional_stats}.

\section{Evaluation}

\subsection{Evaluation Metrics}

We compare author-assigned keyphrases with those extracted by the models, first lemmatizing both sets to account for word variations. Two common matching techniques are used: \textit{Exact matching}, where the extracted keyphrase must exactly match an author-assigned keyphrase, and \textit{Partial matching}, where a match is counted if any fragment of the extracted keyphrase overlaps with a fragment of a golden keyphrase.

Evaluations are performed using $F1$ score at a fixed number \(k\) of keyphrases (denoted as \(F1@k\)), with recall capped at 1 when the number of matches exceeds the golden set (if $k > |\text{golden set}|$). The final $F1$ score is the micro-average across all test documents. We note that both metrics were designed primarily for English, where inflectional variation is limited. In morphologically rich languages, exact matching penalizes correct concepts expressed in a different case or number, while partial matching may over-reward coincidental token overlap. Reporting both metrics and analyzing their divergence therefore provides a more informative picture of model quality than either metric alone---a practice we recommend for any keyphrase evaluation on morphologically rich languages.

\section{Results}

\subsection{Comparing Baseline Models}

\begin{figure}[htbp]
    \centering
    \includegraphics[width=.45\textwidth]{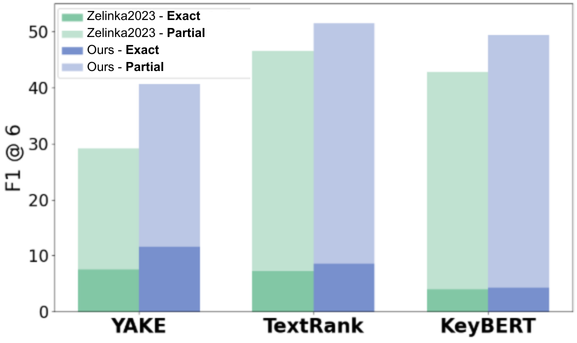}
\caption{The $F1@6$ score for exact (stronger color) and partial (lighter color) matches is shown for the baseline models, comparing Zelinka and SlovKE.}
    \label{fig:bsln_f1} 
\end{figure}

We compared models from three categories---statistical (YAKE), graph-based (TextRank), and pre-trained (KeyBERT)---using the same configuration as in prior work (e.g., $F1@6$, uni- and bi-gram extraction). Figure~\ref{fig:bsln_f1} provides an overview of the comparison with the Zelinka dataset.

\begin{figure}[htbp]
    \centering
    \includegraphics[width=.45\textwidth]{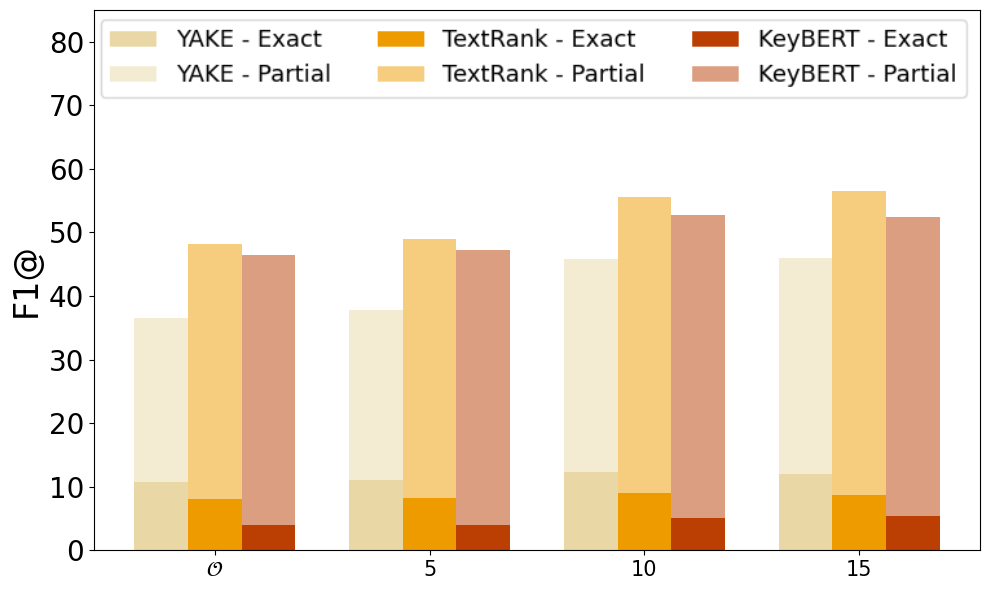}
\caption{F1 score of exact and partial match for baseline models using different $k$ values, where $\mathcal{O}$ denotes $k = | \text{golden set} |$.}
    \label{fig:bsln_f1_k} 
\end{figure}

The evaluations (Table \ref{tab:exact_matching}; Table \ref{tab:partial_matching}) showed slight improvements in exact matching, with YAKE potentially benefiting from the removal of noisy text during the data cleaning stage. This was because we cleaned the abstracts by removing words at the beginning of the text, such as \textit{ABSTRACT} or \textit{Bachelor Thesis}. Due to its underlying scoring mechanism, YAKE assigns high importance to such tokens. For partial matching, despite the notable improvement, the ranking of top-performing models remains consistent. Additional analysis across various values of $k$ (Figure~\ref{fig:bsln_f1_k}) shows that YAKE achieves the highest $F1$ score in exact matching, while TextRank performs best in partial matching.

\begin{table}  
    \centering
    \renewcommand{\arraystretch}{1} 
    \setlength{\tabcolsep}{4pt} 

    \begin{tabular}{llcc}
        \toprule
        \multicolumn{4}{c}{\textbf{Exact Matching}} \\
        \midrule
        \textbf{Model} & \textbf{Metric} & \textbf{SlovKE} & \textbf{Zelinka} \\
        \midrule
        \multirow{3}{*}{YAKE} 
        & Precision  & \textbf{10.4} & 6.8  \\
        & Recall     & \textbf{13.2} & 8.6  \\
        & $F1@6$     & \textbf{11.6} & 7.5  \\
        \midrule
        \multirow{3}{*}{TextRank} 
        & Precision  & \textbf{7.7} & 6.5  \\
        & Recall     & \textbf{9.8} & 8.3  \\
        & $F1@6$     & \textbf{8.6} & 7.2  \\
        \midrule
        \multirow{3}{*}{KeyBERT} 
        & Precision  & \textbf{4.0} & 3.8  \\
        & Recall     & \textbf{4.7} & 4.4  \\
        & $F1@6$     & \textbf{4.3} & 4.0  \\
        \bottomrule
    \end{tabular}
    \caption{Comparison of keyphrase extraction models using \textit{Exact Matching}.}
    \label{tab:exact_matching}
    \centering
    \renewcommand{\arraystretch}{1} 
    \setlength{\tabcolsep}{4pt} 

    \begin{tabular}{llcc}
        \toprule
        \multicolumn{4}{c}{\textbf{Partial Matching}} \\
        \midrule
        \textbf{Model} & \textbf{Metric} & \textbf{SlovKE} & \textbf{Zelinka} \\
        \midrule
        \multirow{3}{*}{YAKE} 
        & Precision  & \textbf{36.9} & 26.6  \\
        & Recall     & \textbf{46.0} & 33.2  \\
        & $F1@6$     & \textbf{40.6} & 29.1  \\
        \midrule
        \multirow{3}{*}{TextRank} 
        & Precision  & \textbf{48.1} & 43.9  \\
        & Recall     & \textbf{56.3} & 50.6  \\
        & $F1@6$     & \textbf{51.5} & 46.5  \\
        \midrule
        \multirow{3}{*}{KeyBERT} 
        & Precision  & \textbf{46.6} & 40.8  \\
        & Recall     & \textbf{53.7} & 46.4  \\
        & $F1@6$     & \textbf{49.4} & 42.8  \\
        \bottomrule
    \end{tabular}
    \caption{Comparison of keyphrase extraction models using \textit{Partial Matching}.}
    \label{tab:partial_matching}
\end{table}

For all $k$, the gap between exact and partial matching is striking: YAKE's exact $F1@6$ of 11.6 vs.\ partial $F1@6$ of 40.6 represents a 29-point spread, and TextRank shows an even larger gap of 43 points (8.6 vs.\ 51.5). We argue that this \emph{exact--partial gap} is a diagnostic metric in its own right: it quantifies the degree to which morphological inflection distorts automated evaluation. In English, where inflection is minimal, the two metrics are far closer. The magnitude of the gap we observe is therefore not merely a Slovak-specific finding---it is predictable for any language with rich nominal and verbal morphology (e.g., Czech, Polish, Finnish, Turkish, Hungarian) and suggests that standard exact-match $F1$ systematically underestimates extractive model performance in such languages. Ideally, we aim not only to achieve a higher $F1$ score for both exact and partial matching but also to narrow the gap between them, ensuring a more reliable evaluation.

\subsection{KeyLLM Results}

To extend our model evaluations, we employed GPT-3.5-turbo with prompts in both Slovak and English, yielding no significant differences, thus we present only results generated using an English prompt (see Appendix~\ref{tab:keyllm_prompt}). 

As GPT-3.5-turbo is costly due to token processing, we initially used embeddings for cost-saving. We employed two Sentence Transformers: \texttt{kinit/slovakbert-sts-stsb} for Slovak, and \texttt{all-MiniLM-L6-v2} for English. This method involved clustering similar documents and selecting one representative for keyphrase extraction, reducing document processing. The model uses a threshold parameter: if the similarity score exceeds the threshold, documents are grouped together; otherwise, they are processed separately. A higher threshold (e.g., 0.9) results in more precise grouping with fewer documents per group, while a lower threshold (e.g., 0.7) reduces costs by clustering more diverse documents, though it may introduce noise.

\begin{figure}[htbp]
    \centering
    \includegraphics[width=.45\textwidth]{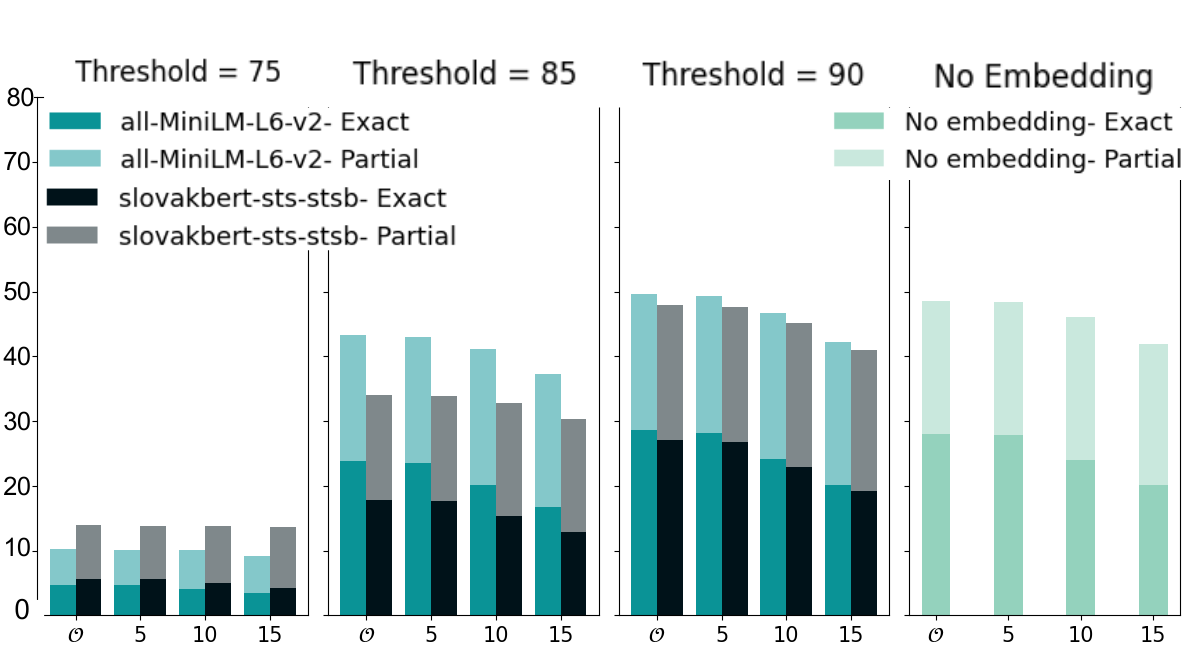}
    \caption{F1 score for KeyLLM with embeddings for thresholds 75, 85, 90, and without embeddings for two Sentence Transformers.}
    \label{fig:keyllm_f1} 
\end{figure}

However, this approach was ineffective with the recommended 0.75 threshold (Figure~\ref{fig:keyllm_f1}). Higher thresholds (0.85) improved results, especially for \texttt{all-MiniLM-L6-v2}, though both models performed similarly at 0.90, with results comparable to keyphrase extraction without embeddings. The computational time and number of processed documents were also similar for the 0.90 threshold or the model without embeddings.

The results suggest that embeddings perform best when small clusters are formed (higher threshold), but for optimal keyphrase extraction, it is better to avoid them. Using the \texttt{all-MiniLM-L6-v2} transformer at thresholds of 0.80–0.85 yields significant improvements over baseline models for exact matching, making it well-suited for large datasets where some accuracy trade-off is acceptable. 

Notably, KeyLLM's exact-match $F1@6$ (${\sim}$15.2) is substantially higher than the best extractive baseline (YAKE, 11.6), while its partial-match $F1@6$ (${\sim}$49.1) is comparable. The resulting exact--partial gap for KeyLLM (${\sim}$34 points) is narrower than that of the extractive models in relative terms: KeyLLM closes roughly 30\% of the gap that YAKE exhibits at $F1@6$. This pattern provides direct evidence that generative models are more robust to morphological inflection than extractive methods, because they can produce canonical lemma forms rather than copying inflected surface tokens. We expect this advantage to generalize to other morphologically rich languages where the same extractive--generative asymmetry holds. However, partial matching showed no significant improvement over the best baselines, suggesting that KeyLLM's gains are concentrated in form normalization rather than in identifying additional relevant concepts. We therefore conducted a further manual evaluation to disentangle these factors.

\subsection{Manual Evaluation}

While standard keyphrase extraction evaluation relies on automatic matching against gold standard keyphrases \citep{kim-etal-2010-semeval}, we additionally performed manual evaluation on a subset of 100 randomly selected documents to better understand the quality of our dataset and the practical utility of extracted keyphrases. This evaluation aimed to assess whether KeyLLM and the YAKE model extract relevant keyphrases beyond exact matches with author-assigned keyphrases. We also tested YAKE with a larger n-gram range (1,3) to examine whether this adjustment improved results.

The manual evaluation was conducted to complement automated 
matching metrics and to better assess the semantic relevance 
of extracted keyphrases. A total of 100 documents were randomly 
sampled from the test set. Each document was evaluated with 
reference to the author-assigned keyphrases and the keyphrases 
extracted by the evaluated models.

For each document, annotators reviewed the abstract and the 
corresponding list of extracted keyphrases. A keyphrase was 
considered correct if it satisfied the following criteria:

\begin{enumerate}
    \item \textbf{Topical relevance:} The keyphrase referred 
    to a central concept explicitly discussed in the abstract.
    \item \textbf{Conceptual adequacy:} The keyphrase 
    represented a meaningful concept rather than a generic 
    modifier (e.g., standalone adjectives were not considered 
    sufficient).
    \item \textbf{Non-redundancy:} When multiple surface 
    variants of the same concept were extracted (e.g., inflected 
    forms or reordered phrases), only the most representative 
    form was retained.
    \item \textbf{Linguistic validity:} Morphological variants 
    (e.g., inflected forms) were accepted as correct if they 
    clearly referred to the same concept as the author-assigned 
    keyphrase.
\end{enumerate}

Two evaluation settings were defined:

\begin{enumerate}
    \item \textbf{Manual Exact}, which counted only keyphrases 
    that matched or were equivalent to the author-assigned 
    keyphrases (including acceptable morphological variants).
    \item \textbf{Manual Extended}, which additionally included 
    contextually relevant keyphrases present in the abstract 
    but omitted from the author-assigned list (e.g., 
    methodologies, named entities, or key concepts discussed 
    but not listed as keyphrases).
\end{enumerate}

The evaluation was primarily conducted by one of the authors,
with a second author independently annotating a subset of 30
documents to assess inter-annotator agreement. Both annotators
were native Slovak speakers with a background in NLP. Agreement
was measured using Cohen's kappa coefficient, yielding
$\kappa = 0.61$, which indicates substantial agreement
according to standard interpretation guidelines
\citep{landis1977measurement}. We note that keyphrase
relevance judgments are inherently subjective---even on
English benchmarks, inter-annotator agreement for keyphrase
annotation rarely exceeds $\kappa = 0.70$
\citep{kim-etal-2010-semeval}---and the morphological
variability of Slovak introduces additional ambiguity in
deciding whether an inflected form constitutes a match.

\subsubsection{Key Observations and Failure Analysis}

\paragraph{Manual vs. Automated Evaluation:} Manual evaluations, particularly for exact matches, consistently outperformed automated methods. While automated algorithms struggled with lexical variation and word ordering, manual evaluation was able to recognize semantically related concepts expressed in different forms. For instance, in a thesis whose topic was obesity, the author-assigned keyphrases covered the core medical condition but omitted concepts that were explicitly discussed in the abstract, such as the psychological consequences of obesity or diagnostic procedures. KeyLLM extracted these as additional keyphrases (e.g., \textit{psychologický problém}, \textit{diagnostika}), which were judged relevant under our Manual Extended criterion despite not appearing in the author-assigned list. These are not matches for \textit{obezita} but rather complementary keyphrases that improve topic coverage.

KeyLLM also outperformed YAKE in root-form handling and contextual understanding. It more reliably converted extracted keyphrases to their base forms and was able to connect words into coherent multi-word keyphrases (e.g., \textit{linguistic characteristics} or \textit{social characteristics}), even when the individual words did not appear contiguously in the text. This ability to produce lemmatized, canonical-form keyphrases without an explicit morphological analyzer is a practical advantage of generative LLMs that extends beyond Slovak: any language where authors assign keyphrases in citation form but text contains inflected forms would benefit similarly. Another notable strength of KeyLLM was its ability to capture distinct keyphrases without redundancy, including named entities such as company names, software tools, and medical systems. In contrast, YAKE frequently extracted semantically similar words or partial fragments of other keyphrases.

\begin{figure}
\centering
\includegraphics[width=.45\textwidth]{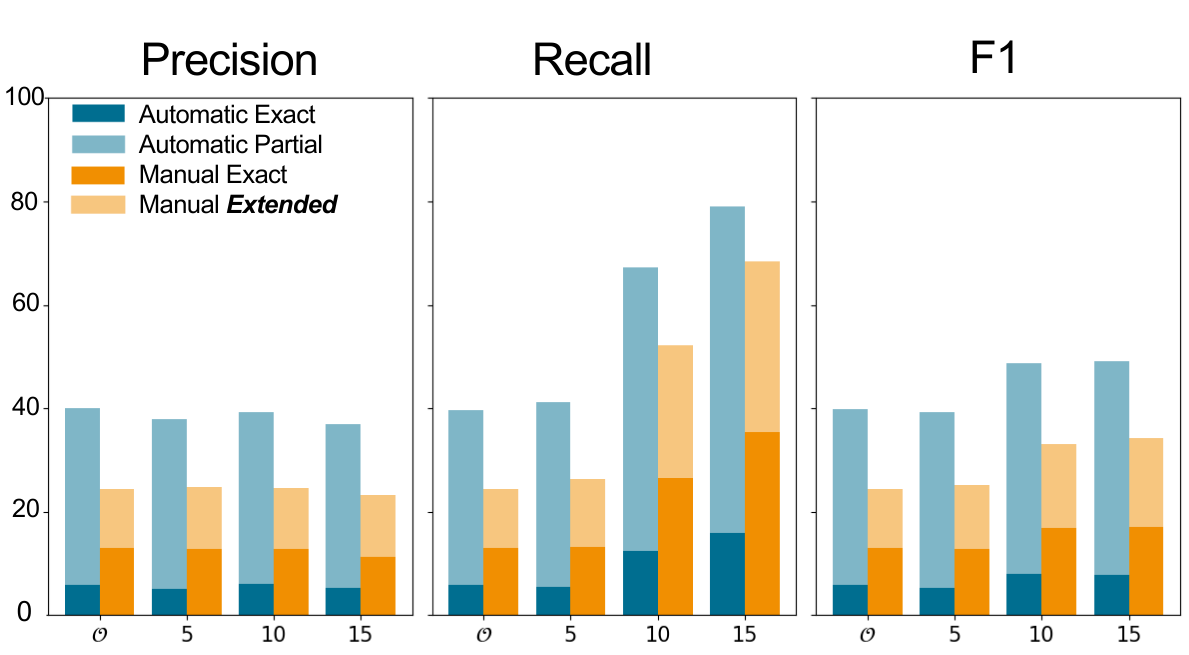}
\caption{Comparison of precision, recall, and F1 score of automatically and manually evaluated YAKE model with n-gram range (1,3)}
\label{fig:man_YAKE3}
\end{figure}

For YAKE, exact-match performance improved under manual evaluation, as annotators could recognize morphologically inflected forms as equivalent to author-assigned keyphrases. However, extended matching did not outperform partial automated matching, since manual evaluation retained only the most representative keyphrase per concept and excluded repetitive surface variations. When increasing the n-gram range to (1,3), YAKE’s manual evaluation scores dropped significantly (Figure~\ref{fig:man_YAKE3}), as the model increasingly prioritized irrelevant trigrams over the document’s main topics.

\subsubsection{Failure Cases}

\paragraph{Morphological mismatch} was the most frequent failure
mode observed for YAKE. Since YAKE extracts keyphrases strictly
in the form in which they appear in the text, Slovak's rich
inflectional morphology regularly caused mismatches with
author-assigned keyphrases even when the correct concept was
identified (e.g., \textit{aktívn\textbf{ej} politik\textbf{y}
trhu} instead of \textit{aktív\textbf{na} politik\textbf{a}
trhu}). This also affected cases where a keyphrase appeared
multiple times in different inflected forms or word orders, and
YAKE selected a surface form that did not match the author's
version. This failure mode is inherent to any extractive method
operating on a morphologically rich language: the same mismatch
pattern would arise in Czech, Polish, Finnish, or Turkish, where
a single lemma can surface in dozens of case- or
agreement-inflected forms. The practical implication is that
extractive keyphrase methods require a lemmatization or
morphological normalization step to be fairly evaluated in
such languages.

\paragraph{Wrong granularity} was particularly prominent for YAKE with the (1,3) n-gram range. The model frequently prioritized irrelevant trigrams that were either surface-level variants of previously extracted keyphrases or generic constructions with little discriminative value (e.g., \textit{analysis of trends}, \textit{trends in data}). This failure mode is partially obscured by automated evaluation, where partial matching counts such variations as correct. Manual evaluation rejected these phrases as redundant, explaining the  gap between manual and automated scores observed with larger n-gram ranges (Figure~\ref{fig:man_YAKE3}).

\paragraph{Semantic redundancy} was another recurring issue for YAKE. The model often occupied multiple extraction slots with surface variants of the same concept (e.g., extracting \textit{rozvoj}, \textit{rozvoj vidieka}, and \textit{vidiecky rozvoj} as separate keyphrases), thereby reducing the effective coverage of distinct document topics.

\paragraph{Unmotivated adjective extraction} represented the primary failure case for KeyLLM. The model frequently extracted standalone adjectives that lacked sufficient topical specificity when not paired with a noun. This behavior also explains the steeper decline in precision with increasing $k$: while highly relevant keyphrases are prioritized early, adjective-only extractions accumulate as more keyphrases are considered.

\section{Conclusion}

This work presents SlovKE, a 227,432-document Slovak keyphrase extraction dataset -- a 25-fold increase over prior work -- created through systematic data cleaning and quality control.

Our evaluation of baseline models reveals that data quality has a direct impact on keyphrase extraction: after cleaning, YAKE's exact $F1@6$ improved from 7.5 to 11.6. More notably, the KeyLLM model (GPT-3.5-turbo) achieved an exact $F1@6$ of approximately 15.2, substantially narrowing the gap between exact and partial matching that characterizes all baseline models. This gap -- with the best exact $F1@6$ at 11.6 versus the best partial $F1@6$ at 51.5 for baseline methods -- reflects a fundamental evaluation challenge for morphologically rich languages, where surface-form variation makes strict matching unreliable and lenient matching overly generous. We argue that this exact--partial divergence is not merely a Slovak-specific artifact but a systematic issue that affects keyphrase evaluation in any language with productive inflectional morphology, including other Slavic, Finno-Ugric, and Turkic languages.

Manual evaluation on 100 documents ($\kappa = 0.61$) confirmed that KeyLLM captures semantically relevant keyphrases that automated metrics systematically miss -- including named entities, methodological terms, and contextually central concepts absent from author-assigned gold standards. These findings suggest that standard automated evaluation underestimates the true quality of keyphrase extraction in morphologically rich languages and that developing morphology-aware evaluation protocols is an important direction for the field.

Beyond its immediate contribution to keyphrase extraction, SlovKE constitutes foundational infrastructure for Slovak NLP: at 227K documents with structured metadata, it is among the largest publicly available collections of Slovak scientific text and can support supervised keyphrase generation, document classification, cross-lingual transfer to typologically similar languages (Czech, Polish, Croatian), and the development of morphology-aware evaluation metrics. The dataset is available at \url{https://huggingface.co/datasets/NaiveNeuron/SlovKE} and the evaluation code at \url{https://github.com/NaiveNeuron/SlovKE}.

\section{Limitations}

\paragraph{Dataset scope.} While our dataset is the largest of its kind for Slovak, it draws exclusively from theses and dissertations registered in the Central Register, which may not capture the full diversity of Slovak academic writing. Extending coverage to journal articles, conference papers, and technical reports would broaden domain representation and is a natural next step for future dataset expansion.

\paragraph{Residual noise.} Despite extensive cleaning, some records may contain formatting inconsistencies or uninformative keyphrases, as neither abstracts nor author-assigned keyphrases underwent expert review. Developing automated quality scoring (e.g., keyphrase informativeness classifiers) could help identify and filter such cases, and our dataset provides a large enough training signal to support such efforts.

\paragraph{Absence of keyphrase rankings.} Our data source does not provide ranked keyphrases, precluding evaluation with rank-sensitive metrics such as MRR \citep{voorhees-tice-2000-trec} or MAP \citep{liu2009encyclopedia}. Crowdsourced or model-assisted ranking annotation over a subset of the dataset would enable these evaluations and constitutes a concrete direction for future work.

\paragraph{Scale of manual evaluation.} Our manual evaluation covered 100 documents with two annotators ($\kappa = 0.61$). While this was sufficient to identify systematic patterns in model behavior, scaling this evaluation -- potentially through crowdsourcing or active-learning-based annotation -- would strengthen the generalizability of our findings across domains and document types.

\paragraph{Model optimization.} The models evaluated in this work were not fine-tuned for Slovak. Given that our dataset provides over 180,000 training documents, supervised fine-tuning of sequence labeling or generative models represents the most direct path to improved performance. In particular, fine-tuning Slovak-specific language models (e.g., SlovakBERT) for keyphrase extraction could yield substantial gains and would directly leverage the resource we provide.

\paragraph{Unsupervised methods only.} We evaluated only unsupervised and prompting-based approaches. Training supervised models on this dataset -- including transformer-based sequence taggers and keyphrase generation models -- is an important next step that could reveal richer structure in the data and substantially improve extraction accuracy.

\section{Data Availability}

To support reproducibility and future research, we publicly release the SlovKE dataset under the Creative Commons Attribution 4.0 International License (CC~BY~4.0) at \url{https://huggingface.co/datasets/NaiveNeuron/SlovKE}. The dataset comprises 227,432 cleaned Slovak scientific abstracts with author-assigned keyphrases, split into training (182,745 documents), validation (22,343), and test (22,344) sets. Each record contains the abstract text, keyphrases, and metadata (university, year, document type). All evaluation scripts and preprocessing code are available at \url{https://github.com/NaiveNeuron/SlovKE} under the MIT License. We encourage the community to use this dataset not only for keyphrase extraction but also for related tasks such as document classification, cross-lingual transfer to other Slavic languages, and the development of morphology-aware evaluation metrics.


\section*{Acknowledgements}

This research was partially funded by the EU NextGenerationEU through the Recovery and Resilience Plan for Slovakia under the project No.\ 09I02-03-V01-00029.

\section{Bibliographical References}\label{sec:reference}

\bibliographystyle{lrec2026-natbib}
\bibliography{literatura, custom}

@inproceedings{song2023survey,
  title = {A Survey on Recent Advances in Keyphrase Extraction from Pre-trained Language Models},
  author = {Song, Mingyang and Feng, Yi and Jing, Liping},
  booktitle = {Findings of the Association for Computational Linguistics: EACL 2023},
  pages = {2108--2119},
  year = {2023},
  address = {Dubrovnik, Croatia},
  publisher = {Association for Computational Linguistics},
  url = {https://aclanthology.org/2023.findings-eacl.161/}
}

@article{wu2022pretrained,
  title = {Pre-trained Language Models for Keyphrase Generation: A Thorough Empirical Study},
  author = {Wu, Di and Ahmad, Wasi Uddin and Chang, Kai-Wei},
  journal = {arXiv preprint arXiv:2212.10233},
  year = {2022}
}

@inproceedings{park2020scientific,
  title = {Scientific Keyphrase Identification and Classification by Pre-Trained Language Models Intermediate Task Transfer Learning},
  author = {Park, Seoyeon and Caragea, Cornelia},
  booktitle = {Proceedings of the 28th International Conference on Computational Linguistics (COLING)},
  pages = {5409--5419},
  year = {2020},
  address = {Barcelona, Spain (Online)},
  publisher = {International Committee on Computational Linguistics},
  url = {https://aclanthology.org/2020.coling-main.472/}
}

@inproceedings{gao2022retrieval,
  title = {Retrieval-Augmented Multilingual Keyphrase Generation with Retriever-Generator Iterative Training},
  author = {Gao, Yifan and Yin, Qingyu and Li, Zheng and Meng, Rui and Zhao, Tong and Yin, Bing and King, Irwin and Lyu, Michael},
  booktitle = {Findings of the Association for Computational Linguistics: NAACL 2022},
  pages = {1233--1246},
  year = {2022},
  address = {Seattle, United States},
  publisher = {Association for Computational Linguistics},
  url = {https://aclanthology.org/2022.findings-naacl.92/}
}

@article{terryn2019termeval,
  title = {In No Uncertain Terms: A Dataset for Monolingual and Multilingual Automatic Term Extraction from Comparable Corpora},
  author = {Terryn, Ayla Rigouts and Hoste, V{\'e}ronique and Lefever, Els},
  journal = {Language Resources and Evaluation},
  volume = {54},
  number = {2},
  pages = {385--418},
  year = {2020},
  publisher = {Springer},
  doi = {10.1007/s10579-019-09453-9}
}

@inproceedings{kulkarni2022maked,
  title = {{MAKED}: A Multi-lingual Automatic Keyword Extraction Dataset},
  author = {Kulkarni, Pruthwik and Singh, Krishnandu and Kulkarni, Manish and Shrivastava, Manish},
  booktitle = {Proceedings of the Thirteenth Language Resources and Evaluation Conference (LREC 2022)},
  pages = {3614--3623},
  year = {2022},
  address = {Marseille, France},
  publisher = {European Language Resources Association},
  url = {https://aclanthology.org/2022.lrec-1.664/}
}

@article{tsarfaty2013parsing,
  title = {Parsing Morphologically Rich Languages: Introduction to the Special Issue},
  author = {Tsarfaty, Reut and Seddah, Djam{\'e} and K{\"u}bler, Sandra and Nivre, Joakim},
  journal = {Computational Linguistics},
  volume = {39},
  number = {1},
  pages = {15--22},
  year = {2013},
  publisher = {MIT Press},
  doi = {10.1162/COLI_a_00133}
}

@inproceedings{sahrawat2020keyphrase,
  title = {Keyphrase Extraction as Sequence Labeling Using Contextualized Embeddings},
  author = {Sahrawat, Dhruva and Mahata, Debanjan and Zhang, Haimin and Kulkarni, Mayank and Sharma, Agniv and Gosangi, Rakesh and Stent, Amanda and Kumar, Yaman and Shah, Rajiv and Zimmermann, Roger},
  booktitle = {Advances in Information Retrieval: 42nd European Conference on IR Research, ECIR 2020},
  pages = {328--335},
  year = {2020},
  address = {Lisbon, Portugal},
  publisher = {Springer},
  doi = {10.1007/978-3-030-45442-5_41}
}

@inproceedings{salaun2024europa,
  title = {{EUROPA}: A Legal Multilingual Keyphrase Generation Dataset},
  author = {Sala{\"u}n, Olivier and Piedboeuf, Fr{\'e}d{\'e}ric and Le Berre, Guillaume and Alfonso-Hermelo, David and Langlais, Philippe},
  booktitle = {Proceedings of the 62nd Annual Meeting of the Association for Computational Linguistics (Volume 1: Long Papers)},
  pages = {12718--12736},
  year = {2024},
  address = {Bangkok, Thailand},
  publisher = {Association for Computational Linguistics},
  url = {https://aclanthology.org/2024.acl-long.687/},
  doi = {10.18653/v1/2024.acl-long.687}
}

@article{landis1977measurement,
  title = {The Measurement of Observer Agreement for Categorical Data},
  author = {Landis, J. Richard and Koch, Gary G.},
  journal = {Biometrics},
  volume = {33},
  number = {1},
  pages = {159--174},
  year = {1977},
  publisher = {International Biometric Society},
  doi = {10.2307/2529310}
}

@preamble{ "\newcommand{\noopsort}[1]{} "
	# "\newcommand{\printfirst}[2]{#1} "
	# "\newcommand{\singleletter}[1]{#1} "
	# "\newcommand{\switchargs}[2]{#2#1} " }

@article{meng2017deep,
  title={Deep keyphrase generation},
  author={Meng, Rui and Zhao, Sanqiang and Han, Shuguang and He, Daqing and Brusilovsky, Peter and Chi, Yu},
  journal={arXiv preprint arXiv:1704.06879},
  year={2017}
}

@article{gallina2019kptimes,
  title={KPTimes: A large-scale dataset for keyphrase generation on news documents},
  author={Gallina, Ygor and Boudin, Florian and Daille, Beatrice},
  journal={arXiv preprint arXiv:1911.12559},
  year={2019}
}

@misc{rogers2002google,
  title={The {Google} {PageRank} Algorithm and How It Works},
  author={Rogers, Ian},
  year={2002},
  howpublished={\url{https://www.cs.princeton.edu/~chazelle/courses/BIB/pagerank.htm}}
}

@inproceedings{kim-etal-2010-semeval,
    title = "{S}em{E}val-2010 Task 5 : Automatic Keyphrase Extraction from Scientific Articles",
    author = "Kim, Su Nam  and
      Medelyan, Olena  and
      Kan, Min-Yen  and
      Baldwin, Timothy",
    editor = "Erk, Katrin  and
      Strapparava, Carlo",
    booktitle = "Proceedings of the 5th International Workshop on Semantic Evaluation",
    month = jul,
    year = "2010",
    address = "Uppsala, Sweden",
    publisher = "Association for Computational Linguistics",
    url = "https://aclanthology.org/S10-1004/",
    pages = "21--26"
}

@inproceedings{hulth2003improved,
  title={Improved automatic keyword extraction given more linguistic knowledge},
  author={Hulth, Anette},
  booktitle={Proceedings of the 2003 conference on Empirical methods in natural language processing},
  pages={216--223},
  year={2003}
}

@inproceedings{pikuliak-etal-2022-slovakbert,
    title = "{S}lovak{BERT}: {S}lovak Masked Language Model",
    author = "Pikuliak, Mat{\'u}{\v{s}}  and
      Grivalsk{\'y}, {\v{S}}tefan  and
      Kon{\^o}pka, Martin  and
      Bl{\v{s}}t{\'a}k, Miroslav  and
      Tamajka, Martin  and
      Bachrat{\'y}, Viktor  and
      Simko, Marian  and
      Bal{\'a}{\v{z}}ik, Pavol  and
      Trnka, Michal  and
      Uhl{\'a}rik, Filip",
    booktitle = "Findings of the Association for Computational Linguistics: EMNLP 2022",
    month = dec,
    year = "2022",
    address = "Abu Dhabi, United Arab Emirates",
    publisher = "Association for Computational Linguistics",
    url = "https://aclanthology.org/2022.findings-emnlp.530",
    pages = "7156--7168",
}

@inproceedings{hasan-ng-2014-automatic,
    title = "Automatic Keyphrase Extraction: A Survey of the State of the Art",
    author = "Hasan, Kazi Saidul  and
      Ng, Vincent",
    editor = "Toutanova, Kristina  and
      Wu, Hua",
    booktitle = "Proceedings of the 52nd Annual Meeting of the Association for Computational Linguistics (Volume 1: Long Papers)",
    month = jun,
    year = "2014",
    address = "Baltimore, Maryland",
    publisher = "Association for Computational Linguistics",
    url = "https://aclanthology.org/P14-1119",
    doi = "10.3115/v1/P14-1119",
    pages = "1262--1273",
}

@inproceedings{under-resourced-lang,
  author       = {Slobodan Beliga and
                  Sanda Martincic{-}Ipsic},
  title        = {Network-Enabled Keyword Extraction for Under-Resourced Languages},
  booktitle    = {Semantic Keyword-Based Search on Structured Data Sources - {COST}
                  Action {IC1302} Second International {KEYSTONE} Conference, {IKC}
                  2016, Cluj-Napoca, Romania, September 8-9, 2016, Revised Selected
                  Papers},
  series       = {Lecture Notes in Computer Science},
  volume       = {10151},
  pages        = {124--135},
  year         = {2016},
  doi          = {10.1007/978-3-319-53640-8\_11}
}

@mastersthesis{Zelinka2023,
    title        = {Keyphrase Extraction from Slovak
Scientific Documents},
    author        = {Andrej Zelinka},
    year          = 2023,
    address      = {Bratislava, Slovakia},
    note         = {Available at \url{https://opac.crzp.sk/?fn=detailBiblioForm&sid=C03CB0FBA6588971DD3A2BD1B6BF}},
    school       = {Comenius University, Bratislava},
    type={Bachelor's Thesis}
}

@inproceedings{varga2022keyphrase,
  author       = {D{\'{a}}vid Varga and
                  Simon Horv{\'{a}}t and
                  Zolt{\'{a}}n Szopl{\'{a}}k and
                  L'ubom{\'{\i}}r Antoni and
                  Stanislav Krajci and
                  Peter Gursk{\'{y}} and
                  Laura Bachn{\'{a}}kov{\'{a}} R{\'{o}}zenfeldov{\'{a}}},
  title        = {Keyphrase extraction from Slovak court decisions},
  booktitle    = {Proceedings of the 22nd Conference Information Technologies - Applications
                  and Theory {(ITAT} 2022)},
  series       = {{CEUR} Workshop Proceedings},
  volume       = {3226},
  pages        = {142--150},
  publisher    = {CEUR-WS.org},
  year         = {2022},
  url          = {https://ceur-ws.org/Vol-3226/paper16.pdf}
}

@article{docekal2022query,
  title={Query-Based keyphrase extraction from long documents},
  author={Docekal, Martin and Smrz, Pavel},
  journal={arXiv preprint arXiv:2205.05391},
  year={2022}
}

@inproceedings{giarelis2021comparative,
  title={A comparative assessment of state-of-the-art methods for multilingual unsupervised keyphrase extraction},
  author={Giarelis, Nikolaos and Kanakaris, Nikos and Karacapilidis, Nikos},
  booktitle={IFIP International Conference on Artificial Intelligence Applications and Innovations},
  pages={635--645},
  year={2021},
  organization={Springer}
}

@inproceedings{pkezik2022keyword,
  title={Keyword extraction from short texts with a text-to-text transfer transformer},
  author={P{\k{e}}zik, Piotr and Miko{\l}ajczyk, Agnieszka and Wawrzy{\'n}ski, Adam and Nito{\'n}, Bart{\l}omiej and Ogrodniczuk, Maciej},
  booktitle={Asian Conference on Intelligent Information and Database Systems},
  pages={530--542},
  year={2022},
  organization={Springer}
}

@article{qi2020stanza,
  title={Stanza: A Python natural language processing toolkit for many human languages},
  author={Qi, Peng and Zhang, Yuhao and Zhang, Yuhui and Bolton, Jason and Manning, Christopher D},
  journal={arXiv preprint arXiv:2003.07082},
  year={2020}
}

@misc{grootendorst2020keybert,
  author       = {Maarten Grootendorst},
  title        = {KeyBERT: Minimal keyword extraction with BERT.},
  year         = 2020,
  publisher    = {Zenodo},
  version      = {v0.3.0},
  doi          = {10.5281/zenodo.4461265},
  url          = {https://doi.org/10.5281/zenodo.4461265}
}

@techreport{radford2018improving,
  title={Improving Language Understanding by Generative Pre-Training},
  author={Radford, Alec and Narasimhan, Karthik and Salimans, Tim and Sutskever, Ilya},
  year={2018},
  institution={OpenAI}
}

@article{achiam2023gpt,
  title={Gpt-4 technical report},
  author={Achiam, Josh and Adler, Steven and Agarwal, Sandhini and Ahmad, Lama and Akkaya, Ilge and Aleman, Florencia Leoni and Almeida, Diogo and Altenschmidt, Janko and Altman, Sam and Anadkat, Shyamal and others},
  journal={arXiv preprint arXiv:2303.08774},
  year={2023}
}

@article{ouyang2022training,
  title={Training language models to follow instructions with human feedback},
  author={Ouyang, Long and Wu, Jeffrey and Jiang, Xu and Almeida, Diogo and Wainwright, Carroll and Mishkin, Pamela and Zhang, Chong and Agarwal, Sandhini and Slama, Katarina and Ray, Alex and others},
  journal={Advances in neural information processing systems},
  volume={35},
  pages={27730--27744},
  year={2022}
}

@article{papagiannopoulou2020review,
  title={A review of keyphrase extraction},
  author={Papagiannopoulou, Eirini and Tsoumakas, Grigorios},
  journal={Wiley Interdisciplinary Reviews: Data Mining and Knowledge Discovery},
  volume={10},
  number={2},
  pages={e1339},
  year={2020},
  publisher={Wiley Online Library}
}

@article{campos2020yake,
  title={YAKE! Keyword extraction from single documents using multiple local features},
  author={Campos, Ricardo and Mangaravite, V{\'\i}tor and Pasquali, Arian and Jorge, Al{\'\i}pio and Nunes, C{\'e}lia and Jatowt, Adam},
  journal={Information Sciences},
  volume={509},
  pages={257--289},
  year={2020},
  publisher={Elsevier}
}

@inproceedings{mihalcea2004textrank,
  title={Textrank: Bringing order into text},
  author={Mihalcea, Rada and Tarau, Paul},
  booktitle={Proceedings of the 2004 conference on empirical methods in natural language processing},
  pages={404--411},
  year={2004}
}

@misc{lingua2021,
  author       = {Lingua},
  title        = {Lingua: Language detection library},
  year         = {2021},
  howpublished = {\url{https://github.com/lingua-project/lingua}},
  note         = {Accessed: 2024-03-04}
}

@inproceedings{voorhees-tice-2000-trec,
    title = "The {TREC}-8 Question Answering Track",
    author = "Voorhees, Ellen M.  and
      Tice, Dawn M.",
    editor = "Gavrilidou, M.  and
      Carayannis, G.  and
      Markantonatou, S.  and
      Piperidis, S.  and
      Stainhauer, G.",
    booktitle = "Proceedings of the Second International Conference on Language Resources and Evaluation ({LREC}`00)",
    month = may,
    year = "2000",
    address = "Athens, Greece",
    publisher = "European Language Resources Association (ELRA)",
    url = "https://aclanthology.org/L00-1018/"
}

@misc{liu2009encyclopedia,
  title={Encyclopedia of database systems},
  author={Liu, Ling},
  year={2009},
  publisher={Springer}
}

@inproceedings{alzaidy2019bi,
  title={Bi-LSTM-CRF sequence labeling for keyphrase extraction from scholarly documents},
  author={Alzaidy, Rabah and Caragea, Cornelia and Giles, C Lee},
  booktitle={The world wide web conference},
  pages={2551--2557},
  year={2019}
}

@inproceedings{basaldella2018bidirectional,
  title={Bidirectional LSTM recurrent neural network for keyphrase extraction},
  author={Basaldella, Marco and Antolli, Elisa and Serra, Giuseppe and Tasso, Carlo},
  booktitle={Italian Research Conference on Digital Libraries},
  pages={180--187},
  year={2018},
  organization={Springer}
}

@inproceedings{nguyen2007keyphrase,
  title={Keyphrase extraction in scientific publications},
  author={Nguyen, Thuy Dung and Kan, Min-Yen},
  booktitle={International conference on Asian digital libraries},
  pages={317--326},
  year={2007},
  organization={Springer}
}

@misc{keyllm,
  author       = {Maarten Grootendorst},
  title        = {KeyLLM: Keyphrase Extraction using Large Language Models},
  year         = {2021},
  howpublished = {\url{https://maartengr.github.io/KeyBERT/guides/keyllm.html}},
  note         = {Accessed: 2024-03-04},
}

\newpage
\appendix

\section{KeyLLM Prompt}
\label{tab:keyllm_prompt}

\begin{table}[h]
\centering
\begin{tabularx}{\linewidth}{lX}
    \toprule
    \textbf{Prompt:} & 
        \texttt{I have the following document:}\\
        & \texttt{[DOCUMENT]}\\
        & \texttt{Based on the information above, extract 15 
        keywords that best describe the topic of the text. 
        Make sure to only extract keywords that appear in the 
        text. Provide the extracted keywords as a list separated 
        by a semicolon. Sort the keywords by their relevance to 
        the text from the most relevant to the least relevant. 
        Do not add any additional text such as "Keywords:" etc.}\\
    \bottomrule
\end{tabularx}
\end{table}

\section{Data Cleaning Details}
\label{sec:cleaning_details}

We identified and removed duplicates, prioritizing documents with complete abstracts and keyphrases. Our final dataset mainly consists of Bachelor and Diploma theses.

\begin{figure}[h]
    \centering
    \includegraphics[width=.45\textwidth]{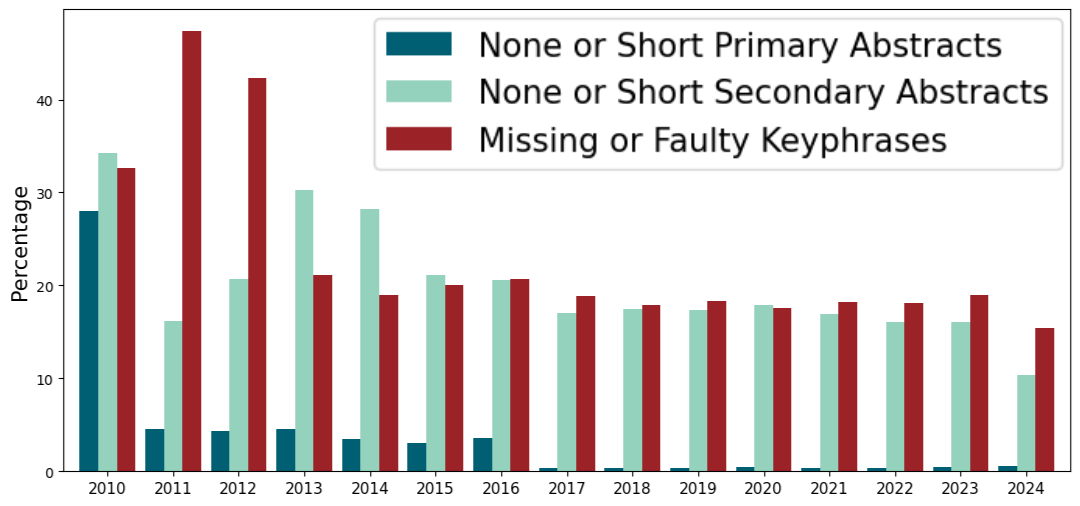}
    \caption{Percentage of missing primary abstracts, secondary abstracts, and keyphrases per year}
    \label{fig:missing_per_year}
\end{figure}

We observed a significant absence of primary abstracts and keyphrases, especially in the first five years (Figure~\ref{fig:missing_per_year}), while missing secondary abstracts and keyphrases persisted across later years.

As shown in Table~\ref{tab:university_missing}, some top-contributing universities exhibit a high proportion of missing keyphrases. We discovered that some universities lack a designated keyphrase field, prompting students to append keyphrases at the end of their abstracts. We attempted to recover these missing keyphrases by identifying and extracting lists appended to abstracts.

\begin{table}[h]
    \centering
    \begin{tabular}{|l|c|c|c|}
        \hline
        \textbf{University} & \textbf{Miss. PA} & \textbf{Miss. SA} & \textbf{Miss. K} \\
               \hline
       {UK BA} & 8.65 & 15.84 & \bfseries{99.94} \\
        {STU BA} & 10.39 & \bfseries{100.00} & 0.01 \\
        {SPU Nitra} & 0.04 & \bfseries{80.71} & 7.91 \\
        {KU RK} & \textbf{56.55} & \textbf{60.65} & 38.29 \\
        {VŠD Sládk.} & 0.83 & 19.33 & \bfseries{86.10} \\
        \hline
    \end{tabular}
    \caption{Top contributing universities with the percentage of missing or short primary abstracts (PA), secondary abstracts (SA), or keyphrases (K). Percentages exceeding 50\% are highlighted in bold.}
    \label{tab:university_missing}
\end{table}

We removed overly short abstracts and keyphrases while also cleaning noisy entries. Many abstracts contained irrelevant metadata prepended to the text, such as:

\parbox[t]{.4\textwidth}{\raggedright
\textcolor{gray}{\textit{ABSTRACT Peter, Doe: Analysis of Various Techniques ...
[Bachelor thesis]. Comenius University in Bratislava. Faculty of Philosophy. Thesis Supervisor: Mgr. John Smith, PhD. Year of Defense: 2019. Number of Pages: 80.}} \\[0.3em]
This thesis explores various techniques for successfully writing an abstract...}

The extra content, typically starting with the author's name or ``ABSTRACT'' and ending with the publication year, was irrelevant and could obscure the abstract. We identified these patterns and trimmed the abstracts accordingly.

We identified abstracts with appended keyphrases (matching `kľúčové slová' or variations), used them to fill missing values, and removed them from the abstract text to prevent model bias.

\begin{figure}[h]
    \centering
    \includegraphics[width=0.45\textwidth]{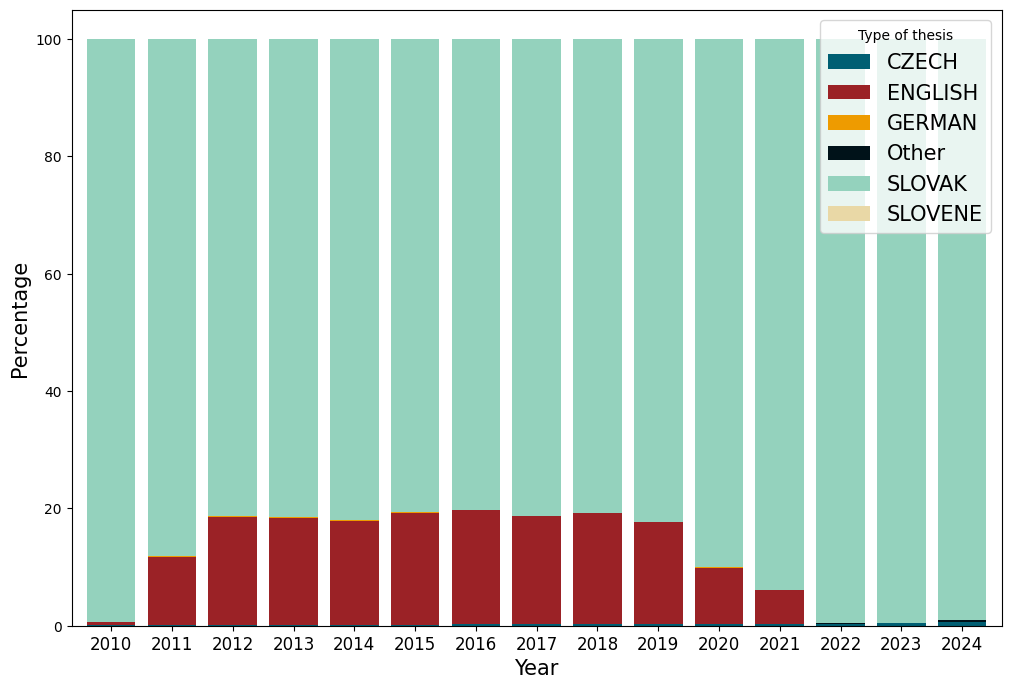}
    \caption{Languages detected in primary abstracts labeled as Slovak}
    \label{fig:lang_first_step}
\end{figure}

We focused on Slovak keyphrase extraction and found that some abstracts labeled as Slovak were actually written in other languages. Using the \texttt{lingua} language detection library, we found that 20\% of Slovak-labeled abstracts (Figure~\ref{fig:lang_first_step}) were actually in English. We implemented multilanguage detection for abstracts with mixed languages, filtering out those with over 40\% non-Slovak content.

Keyphrases were often submitted incorrectly, either as long blocks of text or mixed-language lists. We filtered out keyphrases longer than 300 characters and removed records with insufficient keyphrases or multilingual versions. For keyphrases containing multiple items, we applied multilanguage detection with a 20\% non-Slovak language threshold to exclude English phrases while retaining relevant terminology.

\begin{figure}[h]
     \centering
     \subfigure[Before]{
        \includegraphics[width=0.45\linewidth]{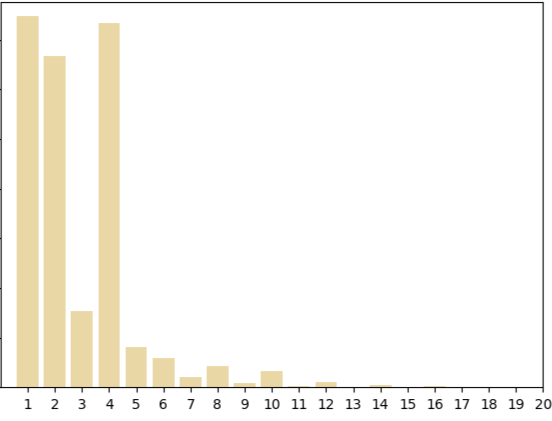}
     }
     \hfill
     \subfigure[After]{
        \includegraphics[width=0.45\linewidth]{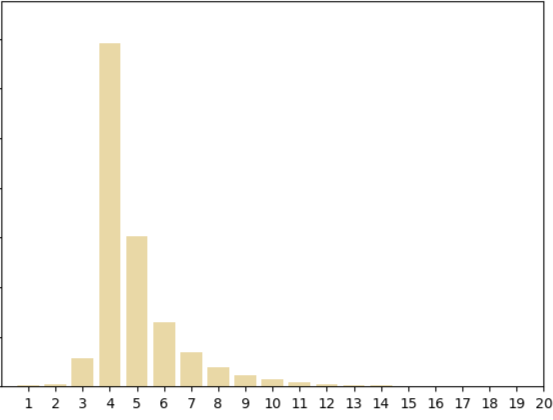}
     }
     \hfill
     \caption{Distribution of the number of keyphrases before and after keyphrase cleaning}
     \label{fig:comparison_keyphr}
\end{figure}

Keyphrase fields often contained lists separated by commas, dots, dashes, or semicolons. We used the \texttt{Stanza} tokenizer to identify words and POS tags, split keyphrases into individual items, removed non-alphabetical characters, and deleted records if any keyphrase exceeded four words. This improved keyphrase length consistency, as shown in Figure~\ref{fig:comparison_keyphr}.

Finally, we filtered abstracts by length (500--2000 characters) and keyphrase count (4--15), retaining documents sufficiently descriptive for model evaluation.

\section{Additional Dataset Statistics}
\label{sec:additional_stats}

\begin{figure}[h]
     \centering
    \includegraphics[width=0.9\linewidth]{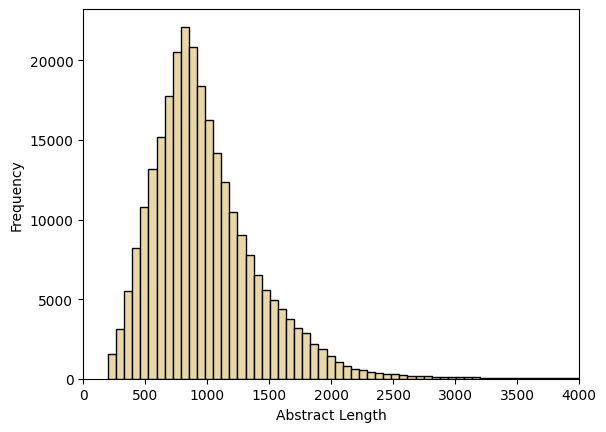}
     \caption{Distribution of abstract lengths before final filtering}
     \label{fig:abstract_len}
\end{figure}

\begin{figure}[h]
    \centering
    \subfigure[Sentences]{
    \includegraphics[width=0.10\textwidth, height=.30\textwidth]{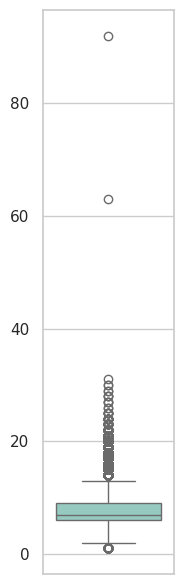}
    }
    \hspace{.03\textwidth}
    \subfigure[Tokens (all tokens (Right), excluding digits and punctuation (Left))]{
    \includegraphics[width=0.25\textwidth]{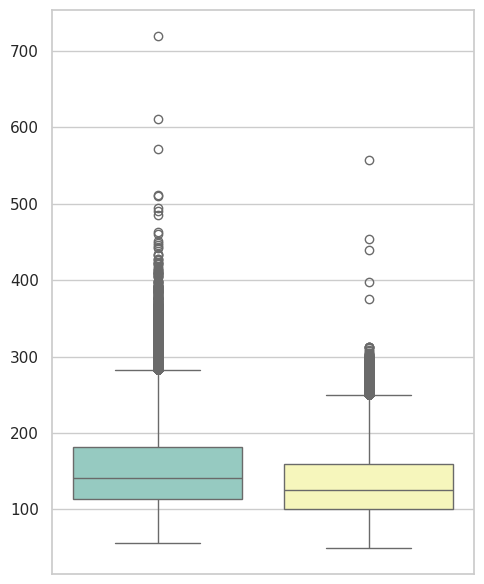}
    }
    \caption{Box plots for the mean number of sentences and tokens in abstracts}
    \label{fig:snet_tok}
\end{figure}

Figure~\ref{fig:abstract_len} shows the distribution of abstract lengths before final filtering, and Figure~\ref{fig:snet_tok} summarizes the number of sentences and tokens per abstract.

\end{document}